\documentclass[10pt,twocolumn,letterpaper]{article}

\usepackage{cvpr}              % To produce the CAMERA-READY version
\usepackage{graphicx}
\usepackage{amsmath}
\usepackage{amssymb}
\usepackage{booktabs}
\usepackage{microtype}
\usepackage{colortbl}  %彩色表格需要加载的宏包
\usepackage{xcolor}
\usepackage{array}   %对表列和表格线的设置需要用到array宏包
\usepackage{pifont}       % \ding{xx}
\usepackage{bbding}       % \Checkmark,\XSolid,... (需要和pifont宏包共同使用)
\usepackage{fontawesome}  % \faCheck,\faTimes
\usepackage{tcolorbox}
\usepackage{multirow}
\usepackage{wrapfig}
\usepackage{tcolorbox}
\newcommand{\cmark}{\ding{51}}%
\usepackage{floatrow}
\newfloatcommand{capbtabbox}{table}[][\FBwidth]
\newcommand{\VarSty}[1]{\textnormal{\ttfamily\color{blue!90!black}#1}\unskip}
\newcommand{\boldparagraph}[1]{\vspace{0.1cm}\noindent{\bf #1.}}

\definecolor{cvprblue}{rgb}{0.21,0.49,0.74}
\definecolor{nvgreen}{RGB}{118, 185, 0}
\usepackage[pagebackref,breaklinks,colorlinks,allcolors=nvgreen]{hyperref}
\usepackage[capitalize]{cleveref}
\def\paperID{13833} % *** Enter the Paper ID here
\def\confName{CVPR}
\def\confYear{2025}

\title{OmniDrive: A Holistic Vision-Language Dataset for Autonomous Driving\\ with Counterfactual Reasoning}

\author{
Shihao Wang$^{1,2*}$,
Zhiding Yu$^{1\dagger}$,
Xiaohui Jiang$^3$,
Shiyi Lan$^1$,
Min Shi$^{1*}$, \\
Nadine Chang$^1$,
Jan Kautz$^1$,
Ying Li$^3$,
Jose M. Alvarez$^1$,
\\[0.2cm]
$^1$NVIDIA~~~~$^2$The Hong Kong Polytechnic University~~~~$^3$Beijing Institute of Technology\\
\href{https://github.com/NVlabs/OmniDrive}{https://github.com/NVlabs/OmniDrive}
}

\begin{document}
\twocolumn[{%
\renewcommand\twocolumn[1][]{#1}%
\maketitle
\vspace{-1.1cm}
\begin{center}
    \includegraphics[width=\linewidth]{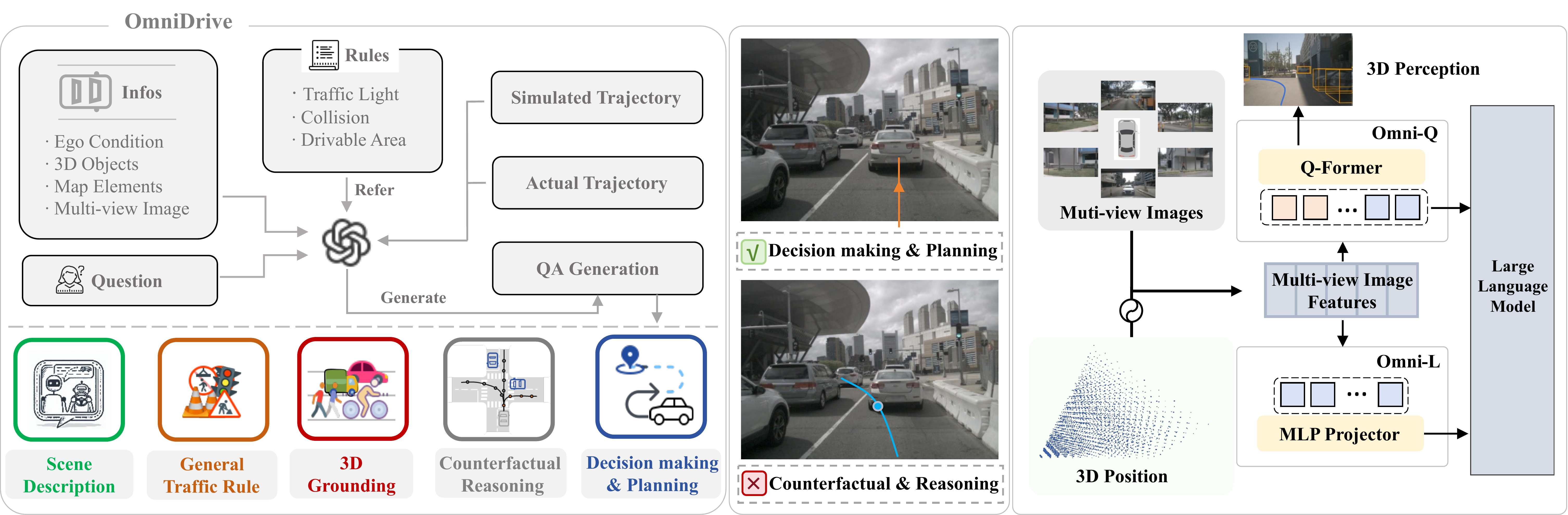}
    \captionof{figure}{OmniDrive is a holistic vision-language dataset for autonomous driving, utilizing counterfactual reasoning to generate high-quality QA data from simulated and actual trajectories. We explore two baseline models: Omni-Q, which designs vision-language models (VLMs) from a 3D perception standpoint, and Omni-L, which builds from VLMs to enhance 3D integration.}
\label{teaser}
\end{center}
}]

\maketitle
% \begin{abstract}
% The advances in multimodal large language models (MLLMs) have led to growing interests in LLM-based autonomous driving to leverage their strong reasoning capabilities. However, capitalizing on MLLMs' strong reasoning capabilities for improved planning behavior is challenging since it requires full 3D situational awareness beyond 2D reasoning. To address this challenge, our work proposes OmniDrive, a holistic framework for strong alignment between agent models and 3D driving tasks. Our framework starts with a novel 3D MLLM architecture that uses sparse queries to lift and compress visual representations into 3D before feeding them into an LLM. This query-based representation allows us to jointly encode dynamic objects and static map elements (e.g., traffic lanes), providing a condensed world model for perception-action alignment in 3D. We further propose a new benchmark with comprehensive visual question-answering (VQA) tasks, including scene description, traffic regulation, 3D grounding, counterfactual reasoning, decision making and planning. Extensive studies show the excellent reasoning and planning capabilities of OmniDrive in complex 3D scenes.
% \end{abstract}

\begin{abstract}
The advances in vision-language models (VLMs) have led to a growing interest in autonomous driving to leverage their strong reasoning capabilities. However, extending these capabilities from 2D to full 3D understanding is crucial for real-world applications. To address this challenge, we propose OmniDrive, a holistic vision-language dataset that aligns agent models with 3D driving tasks through counterfactual reasoning. This approach enhances decision-making by evaluating potential scenarios and their outcomes, similar to human drivers considering alternative actions. Our counterfactual-based synthetic data annotation process generates large-scale, high-quality datasets, providing denser supervision signals that bridge planning trajectories and language-based reasoning. Futher, we explore two advanced OmniDrive-Agent frameworks, namely Omni-L and Omni-Q, to assess the importance of vision-language alignment versus 3D perception, revealing critical insights into designing effective LLM-agents. Significant improvements on the DriveLM Q\&A benchmark and nuScenes open-loop planning demonstrate the effectiveness of our dataset and methods.
\end{abstract}
\vspace{-0.5cm}    
\section{Introduction}
\label{sec:intro}

The recent rapid development of 2D Vision-Language Models (VLMs)~\cite{alayrac2022flamingo,li2023blip,liu2023llava} and their strong reasoning capabilities have led to a stream of applications in end-to-end autonomous driving~\cite{sima2023drivelm,xu2023drivegpt4,wang2023drivemlm,nie2023reason2drive,chen2023driving}. However, extending capabilities from 2D to 3D understanding is crucial for unlocking potential in real-world applications. Although previous works~\cite{marcu2023lingoqa,tian2024drivevlm} have shown successful applications of LLM-agents in autonomous driving (AD), a holistic and principled framework from dataset to LLM-agent is needed to fully extend VLMs' 2D understanding and reasoning capabilities to 3D geometric and spatial understanding.

\begin{NoHyper}
\def\thefootnote{*}\footnotetext{Work done during an internship at NVIDIA.}
\def\thefootnote{\dag}\footnotetext{Corresponding author: \href{mailto:zhidingy@nvidia.com}{zhidingy@nvidia.com}.}
\end{NoHyper}

Recent drive LLM-agent works feature the importance of datasets~\cite{qian2023nuscenes,marcu2023lingoqa,sima2023drivelm,nie2023reason2drive,ding2024holistic,tian2024drivevlm}. Many are presented as question-answering (Q\&A) datasets to train and benchmark the LLM-agent for either reasoning or planning. Noteably, benchmarks that involve planning~\cite{sima2023drivelm,ding2024holistic,tian2024drivevlm} still resort to using expert trajectories for an open-loop setting on real-world sessions (\eg nuScenes). However, recent studies~\cite{zhai2023rethinking,li2023ego} reveal several limitations of open-loop evaluation: implicit biases towards ego status, overly simple planning scenarios, and easy overfit to expert trajectories. 

\boldparagraph{OmniDrive: VLM Dataset}
Expert driving actions provide only sparse supervision~\cite{chen2024vadv2, li2024hydramdpendtoendmultimodalplanning}, primarily reflecting safe trajectories without delving into the complex decision-making and underlying reasoning processes. Relying solely on this sparse supervision makes it challenging to effectively optimize end-to-end driving models. Counterfactual reasoning involves evaluating potential scenarios and their outcomes, similar to how human drivers consider various possibilities to make safer decisions. Therefore, we combine counterfactual reasoning with the chain-of-thought capabilities of VLMs, as shown in Fig.~\ref{teaser}. This approach creates a more effective connection between planning trajectories and language-based reasoning.

Additionally, we found that using simulated trajectories for counterfactual reasoning efficiently identifies key traffic elements in a scene. This process creates a structured and simplified 3D scene representation, making it easier for GPT to understand 3D scenes and generate more effective 3D driving Q\&A data.
To ensure quality, we utilize a rule-based checklist to assess the consequences of potential trajectories. Based on these results, we design prompts for GPT-4 to generate coherent Q\&A, which helps identify which objects require attention and evaluate outcomes based on trajectories. A human-in-the-loop approach is employed in designing the checklist and prompts, ensuring comprehensive coverage of all scenarios. This methodology ensures that data generation is both reliable and interpretable.

\boldparagraph{Omni-L/Q: LLM-agents}
Designing effective Driving Vision-Language Models~\cite{sima2023drivelm, ding2024holistic, wang2023drivemlm} (VLMs) presents a complex and underexplored challenge. A fundamental question is whether to build upon existing 2D VLMs\cite{liu2023llava, liu2024llavanext} and align them with 3D space, or to integrate current 3D perception stacks~\cite{ZhiqiLi2023BEVFormer, wang2022detr3d, liu2022petr, park2022time} into a vision-language framework. To address this, we explore two promising Large Language Model (LLM) frameworks Omni-L and Omni-Q. The Omni-L utilizes state-of-the-art (SoTA) MLP-projection approach (LLaVA~\cite{liu2023llava,liu2023improvedllava}), which enhances the performance of existing VLMs. The Omni-Q is based on the BEV architecture employed by StreamPETR~\cite{liu2022petr, liu2022petrv2, wang2023exploring}, incorporating Q-Former's~\cite{li2023blip} design to investigate the synergy between LLMs and traditional autonomous driving perception tasks. By addressing crucial considerations in the design of autonomous driving LLM-Agents, we conduct a comprehensive comparison of these paradigms in tasks such as counterfactual reasoning and open-loop planning. Our findings indicate that migrating 2D VLMs to 3D is a more straightforward approach compared to integrating traditional 3D perception stacks into VLMs.

\boldparagraph{Contributions}
We propose OmniDrive, a holistic framework for end-to-end autonomous driving with counterfactual-centric dataset and LLM-agents.
With OmniDrive, 
\begin{enumerate}[label=(\arabic*),topsep=0.25\baselineskip]
\setlength{\itemsep}{.25\baselineskip}%
\setlength{\parskip}{0pt}%
\setlength{\topsep}{0pt}
\setlength{\labelindent}{0pt}%
\item We introduce a counterfactual-based 3D driving Q\&A design pipeline that allows for scalable, high-quality data generation.
\item Models pre-trained on OmniDrive showed significant improvement when tested on DriveLM Q\&A benchmark and nuScenes open-loop planning, demonstrating the effectiveness and quality of our dataset.
\item We explore and compare two advanced frameworks Omni-L and Omni-Q, providing critical insights for designing effective LLM-Agents.
\end{enumerate}

\section{OmniDrive}
\begin{figure*}[t]
\centering
\includegraphics[scale=0.17]{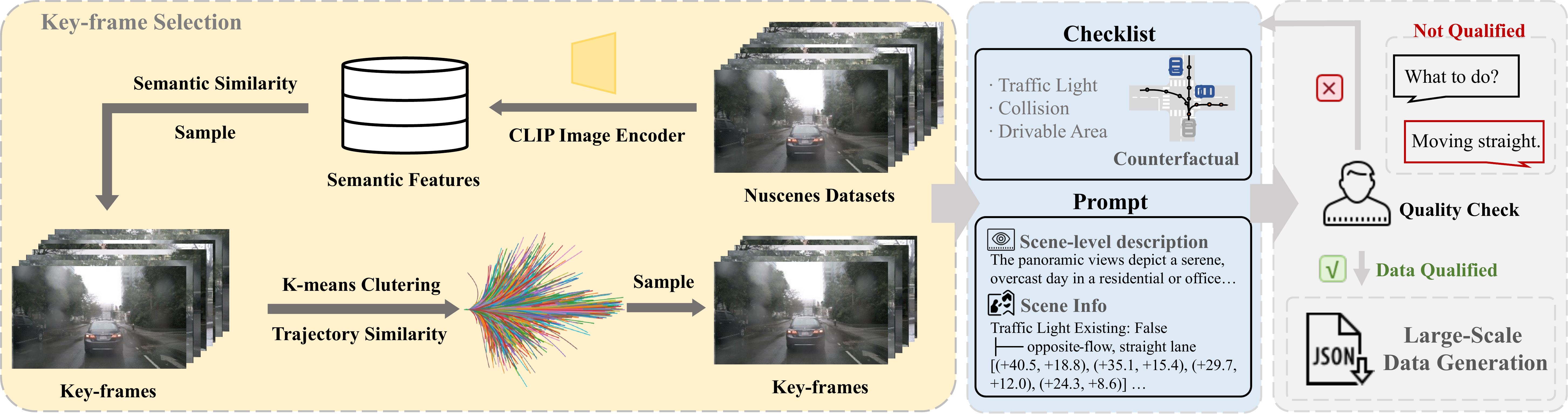}
\caption{The proposed counterfactual-based synthetic Q\&A data generation pipeline integrates semantic key-frame selection, counterfactual-based checklist and prompt design, and human-in-the-loop quality checks to create high-quality Q\&A pairs.}
\label{data_gen}
\end{figure*}

We propose OmniDrive, the counterfactual-based synthetic data on nuScenes~\cite{nuscenes} with high quality Q\&A pairs covering perception, reasoning and planning in 3D domain.

OmniDrive features a human-in-the-loop Q\&A generation pipeline using rule-based checklist and GPT-4. As shown in Fig.~\ref{data_gen}, the data generation process can be divided into: planning-oriented key-frame selection, counterfactual-based checklist and prompt design, human-in-the-loop quality assurance, and large-scale data iteration.

\subsection{Planning-oriented key-frame selection.} 
Autonomous driving datasets often contain significant redundancy, so we focus on selecting representative key-frames to prototype our data processing strategy. We begin by extracting CLIP~\cite{radford2021learning} embeddings from the front view images of the nuScenes~\cite{nuscenes} dataset to capture diverse perceptual elements such as landmarks, traffic lights, and lane markings. Using these embeddings, we apply the K-means algorithm to cluster the data, selecting 20\% of the cluster centers. This ensures that the most semantically representative data is chosen, covering various static and dynamic traffic elements. Next, we further filter the data based on the vehicle's future trajectory. We apply K-means clustering again, this time selecting 200 cluster centers. These centers represent different vehicle dynamics, reflecting driving behaviors such as stopping, moving forward, turning left, turning right, U-turns, accelerating, decelerating, and maintaining constant speed.

This approach effectively compresses the dataset, ensuring that our algorithm design can comprehensively cover all these scenarios. By selecting keyframes, we streamline the process, allowing for more effective rule-based checklist design and prompt iteration. Once the checklist and prompts are verified to cover these scenarios, we iterate on the dataset at a larger scale. This targeted approach ensures that the most relevant data is used for further development.

\subsection{Counterfactual checklist and prompt design.}
Although GPT-4 has powerful ablitity in long-context processing, it cannot effectively understand 3D scenes when directly inputting images, 3D objects, lane markings, and scene definitions. It's difficult to determine the driving status of vehicles or the relationships between traffic elements. This issue becomes more severe as the number of traffic elements in the scene increases. 

To address this, we draw inspiration from counterfactual reasoning and represented the entire scene centered around simulated planned trajectories. In this section, we primarily introduce the types of prompts we input into GPT-4 and how we design checklist on counterfactual principles to enhance the quality of Q\&A data generation.

\noindent\textbf{Simulated trajectories.} We first cluster the driving trajectories from the entire nuScenes dataset. We then classified the cluster centers into categories such as stopping, moving forward, turning left, turning right, making U-turns, accelerating, decelerating, and maintaining a constant speed. In each scene, we simulated these driving behaviors to assess their feasibility. and designed a checklist to determine whether these trajectories violated any traffic rules. 

\noindent\textbf{Counterfactual checklist.} For fixed categories, such as object collisions, road boundary collisions, and running red lights, we use the 3D object detection, centerline and road element topology annotations from nuScenes~\cite{nuscenes} and OpenLane-v2~\cite{wang2024openlane} dataset. We design a rule-based checklist to validate these scenarios. 

However, relying solely on annotated perception elements cannot cover all traffic rules. Therefore, we convert the simulated driving trajectories into high-level decision-making information (this step also involves rule design, such as object and lane assignment, and determining lane-changing behavior, etc.). We then use GPT-4 to analyze the images and assess whether the driving behavior is safe and complies with traffic regulations. We found that using GPT-4 for counterfactual reasoning based on the high-level decision making still achieves good accuracy and interpretability.

\noindent\textbf{Expert trajectory.} We also take the log replay trajectory from nuScenes~\cite{nuscenes} as input prompt. The expert trajectories are classified into different types for high-level decision making. We also identify an object as ``close'', if its minimum distance to the trajectory is smaller than 10 meters in the next 3 seconds. The close objects are then listed below the expert trajectory.

\noindent\textbf{Caption.} To improve the quality of Q\&A data generation, we utilize counterfactual principles to structure and simplify the 3D perception annotations, avoiding the need to input lengthy and unordered scene information. To provide additional contextual information, we also prompt GPT-4 to generate captions, enhancing OCR capabilities and the recognition of open-world object categories.

When both the image and extensive scene information are fed into GPT-4 simultaneously, it tends to overlook details in the image. Therefore, we first prompt GPT-4 to produce a scene description based on multi-view input only. As shown in the top block of Tab.~\ref{tab:teacher}, we stitch the three frontal views and three rear views into two separate images and feed them into GPT-4. We prompt GPT-4 to include the following details: 1) mention weather, time of day, scene type, and other image contents; 2) understand the general direction of each view (\eg the first frontal view being front-left); and 3) avoid mentioning the contents from each view independently, instead describing positions relative to the ego vehicle.

% This approach ensures that GPT-4 focuses on relevant scene details and provides a comprehensive and coherent description, which is crucial for accurate and context-aware decision-making.

\begin{table*}[t!]
\centering
\vspace{-4mm}
\begin{minipage}{0.85\columnwidth}
\centering
\begin{tcolorbox} 
\centering
\scalebox{0.85}{
\begin{tabular}{p{1.1\columnwidth} c}
\centering{\includegraphics[height=4.3cm]{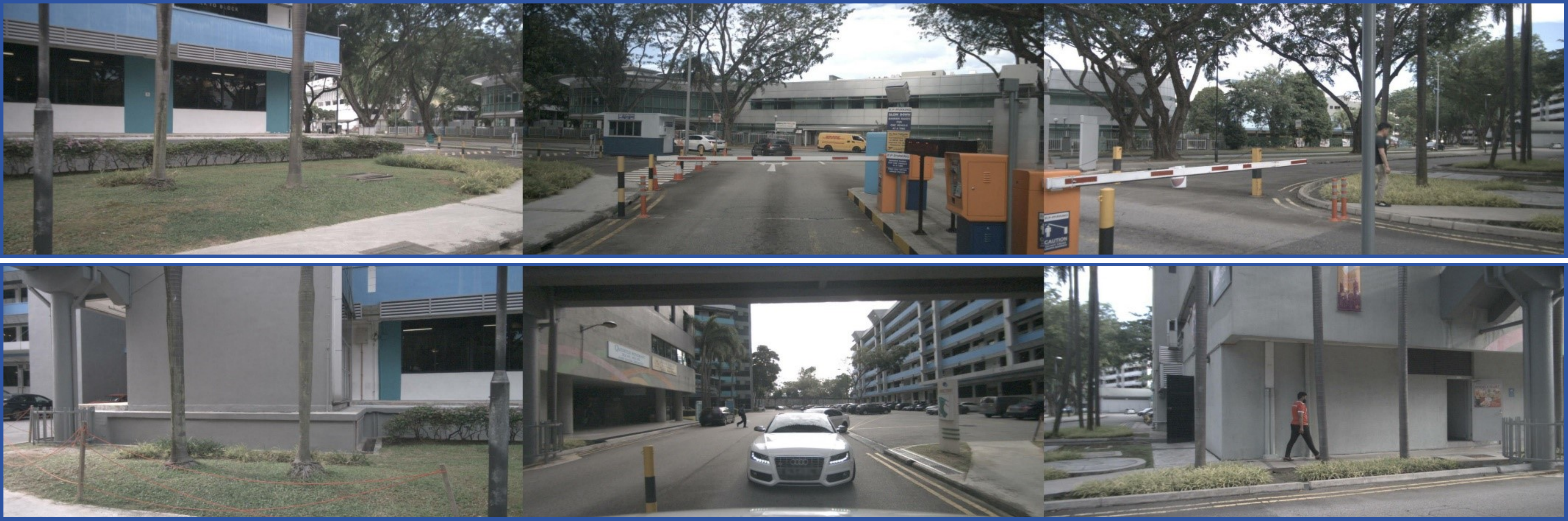}} & \\
\VarSty{ {\bf Prompt type 1: Caption} } & \\
The images depict a daytime setting in a controlled-access area, likely a parking lot or a service entrance of a commercial or industrial facility. On the left, there's a grassy area with trees and a building with blue accents. Moving towards the center, we see a security checkpoint...\textless omitted\textgreater\\

\VarSty{ {\bf Prompt type 2: Simulated decision and trajectory} } & \\
Simulated decision: Moderate Speed, Left Turn \\
Simulated trajectory: [PT, (+4.85, -0.08), (+9.71, -0.22), ..., (+27.42, -0.93)]. \quad

Out of the drivable area\\
\VarSty{ {\bf Prompt type 3: Expert decision and trajectory} } & \\
Expert decision: Moving Slowly, Lane Keeping, Go Straight\\
Expert trajectory: [PT, (+0.76, +0.02), (+1.45, +0.03), ..., (+3.44, +0.12)].\\
Objects need attention: \\
\textbar --- straight lane [(-2.6, +0.5), (+1.2, +0.7), (+5.0, +0.9), (+8.8, +1.0)] \\
\textbar \quad \textbar --- movable\_object.trafficcone at (+8.2, +2.4) \\
\textbar \quad \textbar --- human.pedestrian.moving at (+4.4, -7.2)\\
...\textless omitted\textgreater\\

\hrulefill & \\
\VarSty{ {\bf Conversation type 1: Attention} }  \\
Question: What traffic elements should I be aware of while driving in this area?  \\
Answer: There are movable traffic cones at various points along your path: one directly in front at (+8.2, +2.4), and others in the surrounding lanes, such as those at (+4.7, -6.3)....\textless omitted\textgreater  \\
\VarSty{ {\bf Conversation type 2: Counterfactual reasoning} } \\
Question: If I decide to accelerate and make a left turn, what could be the consequences?\\
Answer:  It appears that you would be turning into the lane designated for allowing from right to left driving. This maneuver could potentially lead to a collision with oncoming traffic...\textless omitted\textgreater& \\
\VarSty{ {\bf Conversation type 3: Decision making and planning} } \\
Question: What should be my next action given the current driving situation, and why?\\
Answer: The most suitable trajectory to follow would be [PT, (+0.76, +0.02), (+1.45, +0.03),...,(+3.44, +0.12)]. This trajectory ensures that you remain in your current straight lane, avoid any obstacles...\textless omitted\textgreater& \\
\VarSty{ {\bf Other Conversations:} } \\
...\textless omitted\textgreater
\end{tabular}
}
\end{tcolorbox}
\vspace{-2mm}
\caption{An example to illustrate the counterfactual-based data generation pipeline. The top block shows contexts such as captions and boxes used to prompt GPT-4, and the bottom block shows the four types of responses. }
\label{tab:teacher}
\end{minipage}
\end{table*}

\subsection{Q\&A generation.}

In the bottom block of the Tab.~\ref{tab:teacher}, we describe the different types of Q\&A responses obtained by using the above context information:

\noindent\textbf{Scene description.} We directly take caption (prompt type 1 in Tab.~\ref{tab:teacher}) as the answer of scene description.

\noindent\textbf{Attention.} Given the simulated and expert trajectories, run simulation to identify close objects. At the same time, we also allowed GPT-4 to use its own common sense to identify threatening traffic elements.

\noindent\textbf{Counterfactual reasoning.} Given the simulated trajectories, we simulate to check if the trajectories violate the traffic rules, such as run a red light, collision to other objects or the road boundary.

\noindent\textbf{Decision making and planning.} We present the high-level decision making as well as the expert trajectory and use GPT-4 to reason why this trajectory is safe, given the previous prompt and response information as context.

\noindent\textbf{General conversation.} We also prompt GPT-4 with generating multi-turn dialogues based on caption information and image content, involving the object countings, color, relative position, and OCR-type tasks. We found that this approach helps improve the model's recognition of long-tail objects.

We design checklists on selected keyframes, followed by prompt design and Q\&A generation. We manually verify the quality of the Q\&A generated from these data. Once our design meets the generalization requirements, we initiate large-scale data generation. This process involves human-in-the-loop quality assurance and large-scale data iteration.
\section{OmniDrive-agent}
\begin{figure*}[t]
\vspace{-6mm}
\centering
\includegraphics[scale=0.15]{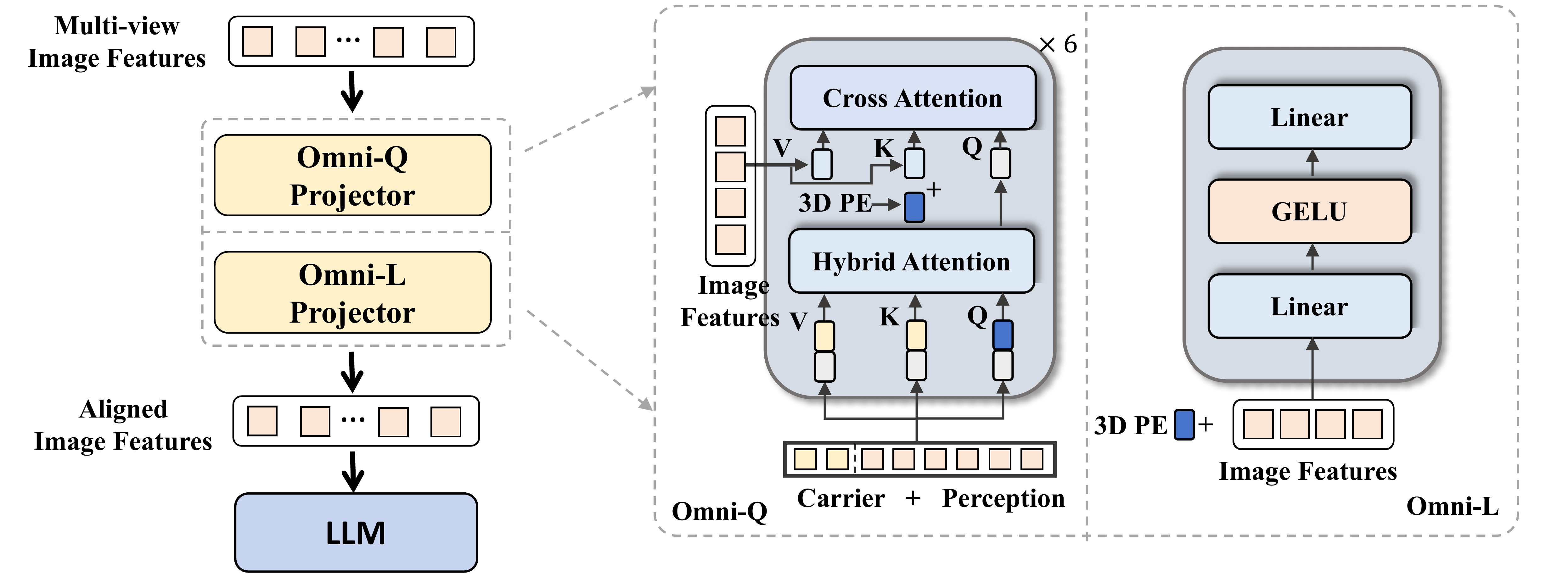}
\caption{Overall pipeline of Omni-L and Omni-Q. The Omni-L. Omni-L follows the 2D VLM design of LLAVA, introducing 3D positional encoding and using MLP layers for vision-language alignment. Omni-Q is based on 3D BEV perception, aligning its architecture with the Q-Former design.}
\label{model_architecture}
\end{figure*}
Designing effective Driving VLMs is a complex challenge. We consider two primary design approaches:

To explore these approaches, we propose two frameworks: Omni-L and Omni-Q. Omni-L leverages the MLP from LLaVA~\cite{liu2023llava} to align multi-view image features to language embedding. Omni-Q is inspired by the BEV architecture of StreamPETR~\cite{wang2023exploring} and incorporates Q-Former's~\cite{li2023blip} design to build the interaction between LLMs and traditional autonomous driving tasks.

Both Omni-L/Q use a shared visual encoder to extract multi-view image features $F_m\in\mathbb{R}^{N \times C \times H \times W}$. The extracted features are combined with the positional encoding 
$P_m$ and then fed into the projector. The visual features are aligned with the text in the projector and then fed into the large language model for text generation tasks.
The main difference between Omni-L and Omni-Q lies in the design of the projector. One prioritizes vision-language alignment, while the other focuses on 3D perception tasks.

%-------------------------------------------------------------------------

\subsection{Omni-Q}
The Transformer decoder in Q-Former~\cite{li2023blip} and the sparse query-based 3D perception models, represented by StreamPETR~\cite{wang2023exploring}, share highly similar architecture designs. To enhance the localization abilities of the VLMs, we consider introducing the design of 3D position encoding $P_m$ and the supervision of the query-based perception models to the training of VLMs.
As shown in Fig.~\ref{model_architecture},  in QFormer, we initialize the detection queries and carrier queries and perform self-attention to exchange their information, which can be summarized by the following formula:
\begin{equation}
    \label{Q-former3d_self}
    \begin{aligned}
    (Q, K, V) = (\textbf{[}Q_{c}, {Q}_{d}\textbf{]},  \textbf{[}Q_{c}, {Q}_{d}\textbf{]}, \textbf{[}Q_{c}, {Q}_{d}\textbf{]}), \\
    \Tilde{Q} = \text{Multi-head Attention}(Q, K, V).
    \end{aligned}
\end{equation}
$\textbf{[}\cdot\textbf{]}$ is the concatenation operation. For simplicity, we omit the position encoding. Then these queries collect information from multi-view images via:
\begin{equation}
    \label{Q-former3d_cross}
    \begin{aligned}
    (Q, K, V) = (\textbf{[}Q_{c}, {Q}_{d}\textbf{]},  P_m + F_m, F_m), \\
    \Tilde{Q} = \text{Multi-head Attention}(Q, K, V).
    \end{aligned}
\end{equation}
After that, the perception queries $Q_{d}$ are used to predict the categories and coordinates of the foreground elements. The carrier queries $Q_{c}$ are sent to a MLP to align with the dimension of LLM tokens (\eg 4096 dimensions in LLaMA2-7B~\cite{touvron2023llama}) and further used for text generation following LLaVA~\cite{liu2023llava}.

In Omni-Q, the carrier queries play the role of the visual-language alignment. Additionally, this design enables carrier queries to leverage the geometric priors provided by the 3D position encoding, while also allowing them to leverage query-based representations acquired through the 3D perception tasks.

\subsection{Omni-L}
Omni-L follows the design of LLaVA~\cite{liu2023llava}, utilizing a simple MLP for aligning the visual-language embedding space. We extend LLaVA's single image input to multiple images, flattening the multi-view image features $F_m$ and feeding them into the large language model. To distinguish different viewpoints, we add 3D position encoding $P_m$ to each image patch. However, for training stability, these position encoding weights are initialized to zero.

\subsection{Training strategy}
\label{Training Strategy}
The training of Omni-L/Q comprises two stages: 2D-Pretraining and 3D-Finetuning. In the initial stage, we pretrain the VLMs on 2D image tasks to initialize the Q-Former/MLP Projector. Following this, the model is fine-tuned on 3D-related driving tasks (\eg motion planning, coutner-factual reasoning, etc.). In both stages, we calculate the 
text generation loss without considering contrasting learning and matching loss for in BLIP-2~\cite{li2023blip}.

%-------------------------------------------------------------------------
\section{Experiment}
\begin{table*}
\begin{center}
\scalebox{0.78}{
\setlength{\tabcolsep}{8pt}
\begin{tabular}{l|cc|cccc|cccc|cccc}
\toprule
\multirow{2}{*}{Method} &
\multicolumn{2}{c|}{Ego Status} &
\multicolumn{4}{c|}{L2 (m) $\downarrow$} & 
\multicolumn{4}{c|}{Collision (\%) $\downarrow$} &
\multicolumn{4}{c}{Intersection (\%) $\downarrow$} \\
 &BEV &Planner& 1s & 2s & 3s &\cellcolor{gray!30}Avg. & 1s & 2s & 3s& 
\cellcolor{gray!30}Avg. & 1s & 2s & 3s &\cellcolor{gray!30}Avg.\\
\midrule
ST-P3 &- &- &\textcolor{gray}{1.59}$^{\text{\textdagger}}$&\textcolor{gray}{2.64}$^{\text{\textdagger}}$&\textcolor{gray}{3.73}$^{\text{\textdagger}}$&\textcolor{gray}
{2.65}$^{\text{\textdagger}}$&\textcolor{gray}{0.69}$^{\text{\textdagger}}$&\textcolor{gray}{3.62}$^{\text{\textdagger}}$&\textcolor{gray}
{8.39}$^{\text{\textdagger}}$& \textcolor{gray}
{4.23}$^{\text{\textdagger}}$& \textcolor{gray}
{2.53}$^{\text{\textdagger}}$&\textcolor{gray}{8.17}$^{\text{\textdagger}}$&\textcolor{gray}
{14.4}$^{\text{\textdagger}}$&\textcolor{gray}{8.37}$^{\text{\textdagger}}$ \\
UniAD &-& -& 0.59 & 1.01 & 1.48 & 1.03& 0.16 & 0.51 & 1.64 &0.77 &0.35&1.46&3.99& 1.93  \\
UniAD &\cmark& \cmark& 0.20 & 0.42 & 0.75& 0.46 & 0.02 & 0.25 & 0.84&0.37 &\textbf{0.20}&\textbf{1.33}&\textbf{3.24}& \textbf{1.59}  \\
\midrule
VAD-Base &-&-& 0.69 & 1.22 & 1.83 &1.25 & 0.06 & 0.68 & 2.52 &1.09 &1.02&3.44&7.00& 3.82  \\
VAD-Base &\cmark& \cmark& 0.17 & 0.34 & 0.60 &0.37 & 0.04 & 0.27 & \textbf{0.67} & 0.33 & 0.21 &2.13&5.06& 2.47  \\
\midrule
Ego-MLP& - & \cmark& 0.15 & 0.32 & 0.59  & 0.35&\textbf{0.00} & {0.27} & 0.85&0.37 &0.27&2.52&6.60& 2.93\\
\midrule
BEV-Planner &- &- & 0.30 & 0.52&0.83 &0.55 & 0.10 & 0.37 & 1.30 &0.59&0.78&3.79& 8.22&4.26\\
BEV-Planner++ &\cmark &\cmark & 0.16 & 0.32& 0.57 & 0.35& \textbf{0.00} & 0.29 & 0.73 &0.34 &0.35&2.62&6.51&3.16 \\
\midrule
\cellcolor[gray]{.9} Omni-Q &-&-& 1.15 & 1.96 & 2.84 & 1.98 & 0.80 & 3.12 & 7.46 & 3.79 & 1.66 & 3.86 & 8.26 &4.59\\
\cellcolor[gray]{.9} Omni-Q++ &\cmark &\cmark & \textbf{0.14} & \textbf{0.29} & \textbf{0.55} & \textbf{0.33} & \textbf{0.00} & \textbf{0.13} & 0.78 & \textbf{0.30} & 0.56 & 2.48 & 5.96 & 3.00 \\
\midrule
\cellcolor[gray]{.9} Omni-L$\ddagger$  &-&-& 1.47 & 2.43 & 3.38 & 2.43 & 0.29 & 2.84 & 6.54 & 3.22 & 1.23 & 3.27 & 7.21 &3.90\\
\cellcolor[gray]{.9} Omni-L++$\ddagger$ & - &\cmark & 0.31 & 0.62 & 1.06 & 0.66 & 0.35 & 2.41 & 0.92 & 0.64 & 2.78 & 2.48 & 5.62 & 3.63 \\
\cellcolor[gray]{.9} Omni-L &-&-& 1.43 & 2.34 & 3.24 & 2.34 & 0.23 & 1.47 & 4.00 & 1.90 & 0.90 & 2.82 & 6.16 &3.29\\
\cellcolor[gray]{.9} Omni-L++ & - &\cmark & 0.15 & 0.36 & 0.70 & 0.40 & 0.06 & 0.27 & 0.72 & 0.35 & 0.49 & 1.99 & 4.86 & 2.45 \\
\bottomrule
\end{tabular}}
\end{center}
\caption{
\textbf{Comparison on nuScenes Open-loop planning.} For a fair comparison, we referred to the reproduced results in BEV-Planner~\cite{li2023ego}. †: The official implementation of ST-P3 (ID-0) utilized partial erroneous ground truth. ${}\ddagger$: The model was trained using only the trajectory prediction task for open-loop planning, without utilizing our generated OmniDrive Q\&A data.}
\label{tab:sota-plan}
 \end{table*}
\subsection{Implementation details}
Our model uses EVA-02-L~\cite{EVA02} as the vision encoder. It applies masked image modeling to distill CLIP~\cite{radford2021learning}, which can extract language-aligned vision features. 

During the 2D pre-training stage, the training data and strategies, including batchsize, learning rate, and optimizer are the same as LLaVA v1.5's~\cite{liu2023improvedllava}.
In the finetuning stage, the model is trained by AdamW~\cite{loshchilov2016sgdr} optimizer with a batch size of 16. The learning rate for the projector is 4e-4, while the visual encoder and the LLM's learning rates are 2e-5. The cosine annealing policy is used for training stability. 

We also explore alternative architectures. The BEV-MLP approach uses LSS method~\cite{philion2020lift, park2022time} to transform perspective features into a BEV feature map. We implement temporal modeling following SOLOFusion~\cite{park2022time}. The BEV features will be consecutively fed into a MLP projector and a LLM. In the following section, both our models -- Omni-Q and Omni-L -- are trained on OmniDrive unless stated otherwise.

\subsection{Dataset \& metrics}
\noindent\textbf{OmniDrive} The proposed OmniDrive dataset involves captioning, open-loop planning and counterfactual reasoning tasks. In this section, we elaborate on how we assess the performance of models on our dataset. For caption-related tasks, such as scene description and the selection of attention objects, we utilize the commonly employed language-based metrics to evaluate the sentence similarity \textbf{CIDEr}~\cite{vedantam2015cider}. Following BEV-Planner~\cite{li2023ego}, \textbf{Collision Rate} and \textbf{Intersection Rate} with the road boundary are adopted to evaluate the performance of Open-loop planning. To evaluate the performance of the counterfactual reasoning, we ask GPT-3.5 to extract keywords based on the predictions. The keywords include `safety,' `collision,', `running a red light,' and `out of the drivable area.' Then we compare extracted keywords with the ground truth to calculate the \textbf{Precision} and \textbf{Recall} for each category of the accident.

\noindent\textbf{DriveLM} The DriveLM~\cite{sima2023drivelm} dataset is designed for end-to-end autonomous driving, featuring Graph Visual Question Answering (GVQA) to handle complex dependencies. It includes 696 scenes from the nuScenes dataset, with 4,072 samples and around 0.3 million image-question pairs. These questions cover perception, prediction, planning, and behavior, helping to fine-tune models and improve performance. DriveLM's evaluation metrics include language metrics like BLEU, ROUGE\_L, and CIDEr for text generation, accuracy for multiple-choice questions, and the ChatGPT Score for open-ended Q\&A. The Match Score assesses the alignment of predicted 2D boxes with ground truth objects. The final score is the weighted combination of GPT Score (0.4), Language Score (0.2), Match Score (0.2), and Accuracy (0.2).

\begin{table}[t]
\centering
\resizebox{\textwidth}{!}{
\setlength{\tabcolsep}{4pt}
\begin{tabular}{lccccccc}
\toprule
\textbf{Dataset} & \textbf{Acc.} & \textbf{CG.} & \textbf{Blue} & \textbf{RL.} & \textbf{Cl.} & \textbf{Mat.} & \textbf{Score} \\ \midrule
DriveLM & 0.60 & 0.65 & 0.50 & 0.71 & 0.07 & 0.36 & 0.53 \\ 
+OmniDrive & 0.70 & 0.65 & 0.52 & 0.73 & 0.13 & 0.37 & 0.56 \\ 
+LLaVA665k & 0.76 & 0.63 & 0.53 & 0.73 & 0.15 & 0.37 & 0.57 \\ 
+Both & \textbf{0.78} & \textbf{0.64} & \textbf{0.54} & \textbf{0.73} & \textbf{0.15} & \textbf{0.37} & \textbf{0.58} \\ 
\bottomrule
\end{tabular}}
\caption{\textbf{The performance of Omni-L on the DriveLM benchmark.} We added pre-training with OmniDrive and LLaVA665k, which significantly improves performance.}
\label{ablation_drivelm}
\end{table}

\begin{table*}[th!]
\centering
\scalebox{0.93}{
\setlength{\tabcolsep}{8pt}
\begin{tabular}{l|c|cc|cc|cc|cc}
\toprule
\multirow{2}{*}{\textbf{Ablation}} & \multirow{2}{*}{\textbf{Exp.}} & \multicolumn{2}{c|}{\textbf{Safe}} & \multicolumn{2}{c|}{\textbf{Red Light}} & \multicolumn{2}{c|}{\textbf{Collision}} & \multicolumn{2}{c}{\textbf{Drivable Area}} \\
\cmidrule{3-10}
& & \textbf{P} & \textbf{R} & \textbf{P} & \textbf{R} & \textbf{P} & \textbf{R} & \textbf{P} & \textbf{R} \\
\midrule
\multirow{3}{*}{\textbf{Architecture}}
&\cellcolor[gray]{.9}Omni-L & \textbf{72.1} & \textbf{58.0} & \textbf{59.2} & \textbf{63.3} & \textbf{34.3} & 71.3 & \textbf{49.1} & \textbf{59.2} \\
&\cellcolor[gray]{.9}Omni-Q & 70.7 & 49.0 & 57.6 & 58.3 & 32.3 & \textbf{72.6} & 48.5 & 58.6 \\
& BEV-MLP & 70.2 & 17.3 & 48.7 & 53.6 & 31.1 & 70.4 & 32.4 & 56.6 \\
\midrule
\textbf{Perception} & No Lane & 67.7 & 57.3 & 58.1 & 59.6 & 31.0 & 56.7 & 47.9 & 56.8 \\
\textbf{Supervision} & No Object \& Lane & 69.0 & 57.8 & 51.3 & 61.2 & 30.0 & 53.2 & 45.3 & 57.1 \\
\bottomrule
\end{tabular}}
\caption{\textbf{Analysis on OmniDrive counterfactual reasoning and open-loop planning (without ego status).} P and R represent Precision and Recall respectively. ``No Object" and ``No Lane" indicate no corresponding 3D perception supervision in Omni-Q.}
\label{ablation_table_planning}
\end{table*}
 
\subsection{Discussion on open-loop planning}
We compare Omni-L and Omni-Q with previous SoTA vision-based planners on nuScenes open-loop planning in Tab.~\ref{tab:sota-plan}. The VLM-based open-loop planning can also achieve comparable performance to SoTA methods when using ego status. However, as mentioned in BEV-planner~\cite{li2023ego}, encoding the ego status significantly improves the metrics across all methods. We found that large language models tend to overfit more easily to ego status. Omni-Q exhibits weaker VLM capabilities, making the overfitting more pronounced compared to Omni-L, as evidenced by a lower L2 error but a higher collision rate. Omni-L performs significantly better than Omni-Q without using ego status. Additionally, if the model is trained solely on trajectory prediction tasks, the distribution modeling capability of the language model can degrade, leading to poor open-loop planning results. However, training with our Q\&A data without ego status effectively mitigates this issue, reducing the collision rate from 3.22\% to 1.90\% and the intersection rate from 3.90\% to 3.29\%. Critically, training Omni-L with both OmniDrive and ego status leads to significant improvements across metrics.

\subsection{Results on DriveLM dataset}
Our dataset, OmniDrive, plays a crucial role in pre-training. Training with only DriveLM dataset leads to an average score of 53\%. However, incorporating our pre-training data boosts performance by 3\%. Additionally, when combined with LLaVA665K pre-training, our data still provides a significant improvement. Although our annotations are highly automated, the OmniDrive dataset maintains high quality through counterfactual-based checklist and human-in-the-loop validation.

\subsection{Planning with counterfactual reasoning} 
We evaluated our OmniDrive counterfactual reasoning tasks in Tab.~\ref{ablation_table_planning}. We observe that Omni-L, designed from a VLM perspective, performs better on average (\eg 72.1\% Precision and 58.0\% Recall in safety tasks) compared to Omni-Q. However, Omni-Q benefits from 3D perception supervision, showing improvements in tasks like collision detection (32.3\% Precision and 72.6\% Recall) compared to those without 3D supervision. The combination of BEV features and MLP results in poorer performance due to pre-training gaps (\eg 17.3\% Recall in safety tasks).

\begin{table*}[th!]
\centering
\setlength{\tabcolsep}{8pt}
\vspace{-4mm}
\begin{tabular}{l|c|cc|c|cc}
\toprule
\multirow{2}{*}{\textbf{Ablation}} &
\multirow{2}{*}{\textbf{Exp.}} &
\multicolumn{2}{c|}{\textbf{Counterfactual}} &
\multicolumn{1}{c|}{\textbf{Language}} &
\multicolumn{2}{c}{\textbf{Open-loop}} \\
&& \textbf{AP} (\%) $\uparrow$ & \textbf{AR} (\%) $\uparrow$ & \textbf{CIDEr} $\uparrow$ & \textbf{Col}(\%) $\downarrow$ & \textbf{Inter}(\%) $\downarrow$\\
\midrule
\multirow{3}{*}{\textbf{Architecture}}
& \cellcolor[gray]{.9} Omni-L & \textbf{53.7} & \textbf{63.0} & \textbf{73.2} & \textbf{1.90} & \textbf{3.29} \\
& \cellcolor[gray]{.9} Omni-Q & 52.3 & 59.6 & 68.6 & 3.79 & 4.59 \\
& BEV-MLP & 45.6 & 49.5 & 59.5 & 4.43 & 8.56 \\
\midrule 
\textbf{Perception}& No Lane & 51.2 & 57.6 & 67.8 & 4.65 & 8.71  \\
\textbf{Supervision} & No Object \& Lane & 48.9 & 57.3  & 67.8 & 6.77 & 8.43 \\
\bottomrule

\end{tabular}
\caption{\textbf{Analysis on nuScenes open-loop planning and OmniDrive counterfactual reasoning and language ability.}}
\label{ablation_table_models}

\end{table*}
\subsection{Ablation study \& analysis}
In Tab.~\ref{ablation_table_models}, we illustrate the performance of different architectures on OmniDrive counterfactual reasoning, language ability, and nuScenes open-loop planning. It shows a positive correlation between language ability and performance in these tasks. Omni-L excels with a counterfactual AP of 53.7\%, AR of 63.0\%, and a Language CIDEr score of 73.2, alongside better open-loop planning metrics (Col 1.90\%, Inter 3.29\%). In contrast, Omni-Q, despite benefiting from 3D perception, has lower results with a counterfactual AP of 52.3\%, AR of 59.6\%, and a CIDEr score of 68.6, due to weaker foundational language skills. This highlights the need for future exploration on aligning traditional 3D perception stacks with language spaces to enhance performance.

\section{Related works}

\subsection{End-to-end autonomous driving}
The objective of end-to-end autonomous driving is to create a fully differentiable system that spans from sensor input to control signals~\cite{zhang2021roach, wu2022trajectoryguided, Prakash2021CVPR}. The current technical road-map is primarily divided into two paths: open-loop autonomous driving and closed-loop autonomous driving.

In the open-loop autonomous driving, the training and evaluation processes are generally conducted on log-replayed real world datasets~\cite{qian2023nuscenes}. Pioneering work UniAD~\cite{hu2023planning} and VAD~\cite{jiang2023vad} integrate modularized design of perception tasks such as object detection, tracking, and semantic segmentation into a unified planning framework. However, Ego-MLP~\cite{zhai2023rethinking} and BEV-Planner~\cite{li2023ego} highlight the limitations of open-loop end-to-end driving benchmarks. In these benchmarks, models may overfit the ego-status information to achieve unreasonably high performance. Researchers are addressing the challenges in open-loop evaluation by introducing closed-loop benchmarks. Recent works, e.g., MILE~\cite{mile2022}, ThinkTwice~\cite{jia2023thinktwice}, VADv2~\cite{chen2024vadv2} leverage CARLA~\cite{dosovitskiy2017carla} as the simulator, which enables the creation of virtual environments with feedback from other agents. Researchers urgently need a reasonable way to evaluate end-to-end autonomous driving systems in the real world. VLM models enable us to perform interpretable analysis and conduct counterfactual reasoning based on a specific trajectory, thereby enhancing the safety redundancy of the agent.

\subsection{Vision-language models}
vision-language models leverage LLMs and various modalities' encoders to successfully bridge the gap between language and other modalities and perform well on multimodal tasks ranging from visual question answer, captioning, and open-world detection. Some VLMs such as CLIP~\cite{radford2021learning} and ALIGN~\cite{jia2021scaling} utilize contrastive learning to create a similar embedding space for both language and vision. More recently, others such as BLIP-2~\cite{li2023blip} explicitly targets multimodal tasks and takes multimodal inputs. For these models, there are two common techniques in order to align language and other input modalities: self-attention and cross-attention. LLaVa~\cite{liu2023llava}, PaLM-E~\cite{driess2023palm}, PaLI~\cite{chen2022pali}, and RT2~\cite{zitkovich2023rt} utilize self-attention for alignment by interleaving or concatenating image and text tokens in fixed sequence lengths. However, self-attention based VLMs are unable to handle high resolution inputs and are unsuitable for autonomous driving with multi-camera high solution images. Conversely, Flamingo~\cite{alayrac2022flamingo}, Qwen-VL~\cite{bai2023qwen}, BLIP-2~\cite{li2023blip}, utilize cross-attention and are able to extract a fixed number of visual tokens regardless of image resolution. Because of this, our model utilizes Qformer architecture from BLIP-2 to handle our high resolution images. 

\subsection{Drive LLM-agents and benchmarks}
\noindent\textbf{Drive LLM-agents.} Given LLM' high performance and ability to align modalities with language, there is a rush to incorporate VLMs/LLMs with autonomous driving (AD). Most AD VLMs methods attempt to create explainable autonomous driving with end-to-end learning. DriveGPT4 leverages LLMs to generate reasons for car actions while also predicting car's next control signals~\cite{xu2023drivegpt4}. Similarly, Drive Anywhere proposes a patch-aligned feature extraction for VLMs that allow it to provide text query-able driving decisions~\cite{wang2023drive}. Other works leverage VLMs through graph-based VQA (DriveLM)~\cite{sima2023drivelm} or chain-of-thought (CoT) design~\cite{tian2024drivevlm, wei2022chain}. They explicitly solve multiple driving tasks alongside typical VLM tasks, such as generating scene description and analysis, prediction, and planning. 

\noindent\textbf{Benchmarks.} To evaluate AD perception and planning, there are various datasets that capture perception, planning, steering, motion data (ONCE~\cite{mao2021one}, nuScenes~\cite{nuscenes}, CARLA~\cite{dosovitskiy2017carla}, Waymo~\cite{Ettinger_2021_ICCV}). However, datasets with more comprehensive lanugage annotations are required to evaluate Drive LLM methods. Datasets focused on perception and tracking include reasoning, or descriptive like captions range from nuScenes-QA~\cite{qian2023nuscenes}, NuPrompt, ~\cite{wu2023language}. HAD and Talk2Car both contain human like advice to best navigate the car~\cite{kim2019CVPR, deruyttere2019talk2car}, while LaMPilot contains labels meant to evaluate transition from human commands to drive action~\cite{ma2023lampilot}. Beyond scene descriptions, DRAMA~\cite{malla2023drama} and Rank2Tell~\cite{sachdeva2024rank2tell} focus on risk object localization. Contrastly, BDD-X, Reason2Drive focus on car explainability by providing reasons behind ego car's action and behavior~\cite{kim2018textual,marcu2023lingoqa, nie2023reason2drive}. LingoQA~\cite{marcu2023lingoqa} has introduced counterfactual questions into the autonomous driving QA dataset. We believe that the interpretability and safety redundancy of autonomous driving in the open-loop setting can be further enhanced by applying counterfactual reasoning to 3D trajectory analysis.
\section{Conclusion}
We present OmniDrive, a holistic framework designed to advance end-to-end autonomous driving using LLM-agents. By introducing a counterfactual-based 3D driving Q\&A pipeline, we enable scalable, high-quality data generation that significantly enhances decision-making capabilities. Pre-trained models on OmniDrive exhibit significant improvements on the DriveLM QA benchmark and nuScenes open-loop planning, underscoring the effectiveness and quality of our dataset. Furthermore, our exploration of two advanced frameworks, Omni-L and Omni-Q, provides valuable insights into the design of effective LLM-agents, highlighting the advantages of vision-language alignment in 3D spaces. These frameworks demonstrate the potential for improved reasoning and perception by integrating language models with 3D environmental understanding.

\noindent\textbf{Limitations.} 
The simulation of counterfactual outcomes, despite moving beyond single trajectories, does not yet consider reactions from other agents. As research on closed-loop planning simulators progresses, we aim to use closed-loop results to enhance effectiveness.

\section{Acknowledgments}
\label{sec:acknowledgments}
The team would like to give special thanks to the NVIDIA TSE Team, including Le An, Chengzhe Xu, Yuchao Jin, and Josh Park, for their exceptional work on the TensorRT deployment of OmniDrive.

{
    \small
    \bibliographystyle{ieeenat_fullname}
    \bibliography{main}

\begin{thebibliography}{56}
\providecommand{\natexlab}[1]{#1}
\providecommand{\url}[1]{\texttt{#1}}
\expandafter\ifx\csname urlstyle\endcsname\relax
  \providecommand{\doi}[1]{doi: #1}\else
  \providecommand{\doi}{doi: \begingroup \urlstyle{rm}\Url}\fi

\bibitem[Alayrac et~al.(2022)Alayrac, Donahue, Luc, Miech, Barr, Hasson, Lenc, Mensch, Millican, Reynolds, et~al.]{alayrac2022flamingo}
Jean-Baptiste Alayrac, Jeff Donahue, Pauline Luc, Antoine Miech, Iain Barr, Yana Hasson, Karel Lenc, Arthur Mensch, Katherine Millican, Malcolm Reynolds, et~al.
\newblock {Flamingo}: a visual language model for few-shot learning.
\newblock In \emph{NeurIPs}, 2022.

\bibitem[Bai et~al.(2023)Bai, Bai, Yang, Wang, Tan, Wang, Lin, Zhou, and Zhou]{bai2023qwen}
Jinze Bai, Shuai Bai, Shusheng Yang, Shijie Wang, Sinan Tan, Peng Wang, Junyang Lin, Chang Zhou, and Jingren Zhou.
\newblock {Qwen-VL}: A frontier large vision-language model with versatile abilities.
\newblock \emph{arXiv:2308.12966}, 2023.

\bibitem[Caesar et~al.(2020)Caesar, Bankiti, Lang, Vora, Liong, Xu, Krishnan, Pan, Baldan, and Beijbom]{nuscenes}
Holger Caesar, Varun Bankiti, Alex~H Lang, Sourabh Vora, Venice~Erin Liong, Qiang Xu, Anush Krishnan, Yu Pan, Giancarlo Baldan, and Oscar Beijbom.
\newblock {nuScenes}: A multimodal dataset for autonomous driving.
\newblock In \emph{CVPR}, 2020.

\bibitem[Chen et~al.(2023)Chen, Sinavski, H{\"u}nermann, Karnsund, Willmott, Birch, Maund, and Shotton]{chen2023driving}
Long Chen, Oleg Sinavski, Jan H{\"u}nermann, Alice Karnsund, Andrew~James Willmott, Danny Birch, Daniel Maund, and Jamie Shotton.
\newblock {Driving with LLMs}: Fusing object-level vector modality for explainable autonomous driving.
\newblock \emph{arXiv:2310.01957}, 2023.

\bibitem[Chen et~al.(2024)Chen, Jiang, Gao, Liao, Xu, Zhang, Huang, Liu, and Wang]{chen2024vadv2}
Shaoyu Chen, Bo Jiang, Hao Gao, Bencheng Liao, Qing Xu, Qian Zhang, Chang Huang, Wenyu Liu, and Xinggang Wang.
\newblock {VADv2}: End-to-end vectorized autonomous driving via probabilistic planning.
\newblock \emph{arXiv:2402.13243}, 2024.

\bibitem[Chen et~al.(2022)Chen, Wang, Changpinyo, Piergiovanni, Padlewski, Salz, Goodman, Grycner, Mustafa, Beyer, et~al.]{chen2022pali}
Xi Chen, Xiao Wang, Soravit Changpinyo, AJ Piergiovanni, Piotr Padlewski, Daniel Salz, Sebastian Goodman, Adam Grycner, Basil Mustafa, Lucas Beyer, et~al.
\newblock {PaLI}: A jointly-scaled multilingual language-image model.
\newblock \emph{arXiv:2209.06794}, 2022.

\bibitem[Deruyttere et~al.(2019)Deruyttere, Vandenhende, Grujicic, Van~Gool, and Moens]{deruyttere2019talk2car}
Thierry Deruyttere, Simon Vandenhende, Dusan Grujicic, Luc Van~Gool, and Marie~Francine Moens.
\newblock {Talk2Car}: Taking control of your self-driving car.
\newblock In \emph{EMNLP-IJCNLP}, 2019.

\bibitem[Ding et~al.(2024)Ding, Han, Xu, Liang, Zhang, and Li]{ding2024holistic}
Xinpeng Ding, Jinahua Han, Hang Xu, Xiaodan Liang, Wei Zhang, and Xiaomeng Li.
\newblock Holistic autonomous driving understanding by bird's-eye-view injected multi-modal large models.
\newblock \emph{arXiv:2401.00988}, 2024.

\bibitem[Dosovitskiy et~al.(2017)Dosovitskiy, Ros, Codevilla, Lopez, and Koltun]{dosovitskiy2017carla}
Alexey Dosovitskiy, German Ros, Felipe Codevilla, Antonio Lopez, and Vladlen Koltun.
\newblock Carla: An open urban driving simulator.
\newblock In \emph{CoRL}, 2017.

\bibitem[Driess et~al.(2023)Driess, Xia, Sajjadi, Lynch, Chowdhery, Ichter, Wahid, Tompson, Vuong, Yu, et~al.]{driess2023palm}
Danny Driess, Fei Xia, Mehdi~SM Sajjadi, Corey Lynch, Aakanksha Chowdhery, Brian Ichter, Ayzaan Wahid, Jonathan Tompson, Quan Vuong, Tianhe Yu, et~al.
\newblock Palm-e: An embodied multimodal language model.
\newblock \emph{arXiv:2303.03378}, 2023.

\bibitem[Ettinger et~al.(2021)Ettinger, Cheng, Caine, Liu, Zhao, Pradhan, Chai, Sapp, Qi, Zhou, Yang, Chouard, Sun, Ngiam, Vasudevan, McCauley, Shlens, and Anguelov]{Ettinger_2021_ICCV}
Scott Ettinger, Shuyang Cheng, Benjamin Caine, Chenxi Liu, Hang Zhao, Sabeek Pradhan, Yuning Chai, Ben Sapp, Charles~R. Qi, Yin Zhou, Zoey Yang, Aur'elien Chouard, Pei Sun, Jiquan Ngiam, Vijay Vasudevan, Alexander McCauley, Jonathon Shlens, and Dragomir Anguelov.
\newblock Large scale interactive motion forecasting for autonomous driving: The waymo open motion dataset.
\newblock In \emph{ICCV}, 2021.

\bibitem[Fang et~al.(2023)Fang, Sun, Wang, Huang, Wang, and Cao]{EVA02}
Yuxin Fang, Quan Sun, Xinggang Wang, Tiejun Huang, Xinlong Wang, and Yue Cao.
\newblock {EVA-02}: A visual representation for neon genesis.
\newblock \emph{arXiv:2303.11331}, 2023.

\bibitem[Hu et~al.(2022)Hu, Corrado, Griffiths, Murez, Gurau, Yeo, Kendall, Cipolla, and Shotton]{mile2022}
Anthony Hu, Gianluca Corrado, Nicolas Griffiths, Zak Murez, Corina Gurau, Hudson Yeo, Alex Kendall, Roberto Cipolla, and Jamie Shotton.
\newblock Model-based imitation learning for urban driving.
\newblock In \emph{NeurIPS}, 2022.

\bibitem[Hu et~al.(2023)Hu, Yang, Chen, Li, Sima, Zhu, Chai, Du, Lin, Wang, et~al.]{hu2023planning}
Yihan Hu, Jiazhi Yang, Li Chen, Keyu Li, Chonghao Sima, Xizhou Zhu, Siqi Chai, Senyao Du, Tianwei Lin, Wenhai Wang, et~al.
\newblock Planning-oriented autonomous driving.
\newblock In \emph{CVPR}, 2023.

\bibitem[Jia et~al.(2021)Jia, Yang, Xia, Chen, Parekh, Pham, Le, Sung, Li, and Duerig]{jia2021scaling}
Chao Jia, Yinfei Yang, Ye Xia, Yi-Ting Chen, Zarana Parekh, Hieu Pham, Quoc Le, Yun-Hsuan Sung, Zhen Li, and Tom Duerig.
\newblock Scaling up visual and vision-language representation learning with noisy text supervision.
\newblock In \emph{ICML}, 2021.

\bibitem[Jia et~al.(2023)Jia, Wu, Chen, Xie, He, Yan, and Li]{jia2023thinktwice}
Xiaosong Jia, Penghao Wu, Li Chen, Jiangwei Xie, Conghui He, Junchi Yan, and Hongyang Li.
\newblock {Think Twice before Driving}: Towards scalable decoders for end-to-end autonomous driving.
\newblock In \emph{CVPR}, 2023.

\bibitem[Jiang et~al.(2023)Jiang, Chen, Xu, Liao, Chen, Zhou, Zhang, Liu, Huang, and Wang]{jiang2023vad}
Bo Jiang, Shaoyu Chen, Qing Xu, Bencheng Liao, Jiajie Chen, Helong Zhou, Qian Zhang, Wenyu Liu, Chang Huang, and Xinggang Wang.
\newblock {VAD}: Vectorized scene representation for efficient autonomous driving.
\newblock \emph{arXiv:2303.12077}, 2023.

\bibitem[Kim et~al.(2018)Kim, Rohrbach, Darrell, Canny, and Akata]{kim2018textual}
Jinkyu Kim, Anna Rohrbach, Trevor Darrell, John Canny, and Zeynep Akata.
\newblock Textual explanations for self-driving vehicles.
\newblock \emph{ECCV}, 2018.

\bibitem[Kim et~al.(2019)Kim, Misu, Chen, Tawari, and Canny]{kim2019CVPR}
Jinkyu Kim, Teruhisa Misu, Yi-Ting Chen, Ashish Tawari, and John Canny.
\newblock Grounding human-to-vehicle advice for self-driving vehicles.
\newblock In \emph{CVPR}, 2019.

\bibitem[Li et~al.(2023{\natexlab{a}})Li, Li, Savarese, and Hoi]{li2023blip}
Junnan Li, Dongxu Li, Silvio Savarese, and Steven Hoi.
\newblock {BLIP-2}: Bootstrapping language-image pre-training with frozen image encoders and large language models.
\newblock In \emph{ICML}, 2023{\natexlab{a}}.

\bibitem[Li et~al.(2023{\natexlab{b}})Li, Deng, Li, Huang, Sima, Geng, Gao, Wang, Li, and Lu]{ZhiqiLi2023BEVFormer}
Zhiqi Li, Hanming Deng, Tianyu Li, Yangyi Huang, Chonghao Sima, Xiangwei Geng, Yulu Gao, Wenhai Wang, Yang Li, and Lewei Lu.
\newblock {BEVFormer++} : Improving bevformer for 3d camera-only object detection: 1st place solution for waymo open dataset challenge 2022.
\newblock 2023{\natexlab{b}}.

\bibitem[Li et~al.(2023{\natexlab{c}})Li, Yu, Lan, Li, Kautz, Lu, and Alvarez]{li2023ego}
Zhiqi Li, Zhiding Yu, Shiyi Lan, Jiahan Li, Jan Kautz, Tong Lu, and Jose~M Alvarez.
\newblock Is ego status all you need for open-loop end-to-end autonomous driving?
\newblock \emph{arXiv:2312.03031}, 2023{\natexlab{c}}.

\bibitem[Li et~al.(2024)Li, Li, Wang, Lan, Yu, Ji, Li, Zhu, Kautz, Wu, Jiang, and Alvarez]{li2024hydramdpendtoendmultimodalplanning}
Zhenxin Li, Kailin Li, Shihao Wang, Shiyi Lan, Zhiding Yu, Yishen Ji, Zhiqi Li, Ziyue Zhu, Jan Kautz, Zuxuan Wu, Yu-Gang Jiang, and Jose~M. Alvarez.
\newblock Hydra-mdp: End-to-end multimodal planning with multi-target hydra-distillation, 2024.

\bibitem[Liu et~al.(2023{\natexlab{a}})Liu, Li, Li, and Lee]{liu2023improvedllava}
Haotian Liu, Chunyuan Li, Yuheng Li, and Yong~Jae Lee.
\newblock Improved baselines with visual instruction tuning.
\newblock \emph{arXiv:2310.03744}, 2023{\natexlab{a}}.

\bibitem[Liu et~al.(2023{\natexlab{b}})Liu, Li, Wu, and Lee]{liu2023llava}
Haotian Liu, Chunyuan Li, Qingyang Wu, and Yong~Jae Lee.
\newblock Visual instruction tuning.
\newblock NeurIPS, 2023{\natexlab{b}}.

\bibitem[Liu et~al.(2024)Liu, Li, Li, Li, Zhang, Shen, and Lee]{liu2024llavanext}
Haotian Liu, Chunyuan Li, Yuheng Li, Bo Li, Yuanhan Zhang, Sheng Shen, and Yong~Jae Lee.
\newblock {LLaVA-NeXT}: Improved reasoning, ocr, and world knowledge, 2024.

\bibitem[Liu et~al.(2022{\natexlab{a}})Liu, Wang, Zhang, and Sun]{liu2022petr}
Yingfei Liu, Tiancai Wang, Xiangyu Zhang, and Jian Sun.
\newblock {PETR}: Position embedding transformation for multi-view 3d object detection.
\newblock \emph{arXiv:2203.05625}, 2022{\natexlab{a}}.

\bibitem[Liu et~al.(2022{\natexlab{b}})Liu, Yan, Jia, Li, Gao, Wang, Zhang, and Sun]{liu2022petrv2}
Yingfei Liu, Junjie Yan, Fan Jia, Shuailin Li, Qi Gao, Tiancai Wang, Xiangyu Zhang, and Jian Sun.
\newblock {PETRv2}: A unified framework for 3d perception from multi-camera images.
\newblock \emph{arXiv:2206.01256}, 2022{\natexlab{b}}.

\bibitem[Loshchilov and Hutter(2016)]{loshchilov2016sgdr}
Ilya Loshchilov and Frank Hutter.
\newblock {SGDR}: Stochastic gradient descent with warm restarts.
\newblock \emph{arXiv:1608.03983}, 2016.

\bibitem[Ma et~al.(2023)Ma, Cui, Cao, Ye, Liu, Lu, Abdelraouf, Gupta, Han, Bera, et~al.]{ma2023lampilot}
Yunsheng Ma, Can Cui, Xu Cao, Wenqian Ye, Peiran Liu, Juanwu Lu, Amr Abdelraouf, Rohit Gupta, Kyungtae Han, Aniket Bera, et~al.
\newblock {LaMPilot}: An open benchmark dataset for autonomous driving with language model programs.
\newblock \emph{arXiv:2312.04372}, 2023.

\bibitem[Malla et~al.(2023)Malla, Choi, Dwivedi, Choi, and Li]{malla2023drama}
Srikanth Malla, Chiho Choi, Isht Dwivedi, Joon~Hee Choi, and Jiachen Li.
\newblock {DRAMA}: Joint risk localization and captioning in driving.
\newblock In \emph{WACV}, 2023.

\bibitem[Mao et~al.(2021)Mao, Niu, Jiang, Liang, Li, Ye, Zhang, Li, Yu, Xu, et~al.]{mao2021one}
Jiageng Mao, Minzhe Niu, Chenhan Jiang, Xiaodan Liang, Yamin Li, Chaoqiang Ye, Wei Zhang, Zhenguo Li, Jie Yu, Chunjing Xu, et~al.
\newblock One million scenes for autonomous driving: Once dataset.
\newblock 2021.

\bibitem[Marcu et~al.(2023)Marcu, Chen, H{\"u}nermann, Karnsund, Hanotte, Chidananda, Nair, Badrinarayanan, Kendall, Shotton, et~al.]{marcu2023lingoqa}
Ana-Maria Marcu, Long Chen, Jan H{\"u}nermann, Alice Karnsund, Benoit Hanotte, Prajwal Chidananda, Saurabh Nair, Vijay Badrinarayanan, Alex Kendall, Jamie Shotton, et~al.
\newblock {LingoQA}: Video question answering for autonomous driving.
\newblock \emph{arXiv:2312.14115}, 2023.

\bibitem[Nie et~al.(2023)Nie, Peng, Wang, Cai, Han, Xu, and Zhang]{nie2023reason2drive}
Ming Nie, Renyuan Peng, Chunwei Wang, Xinyue Cai, Jianhua Han, Hang Xu, and Li Zhang.
\newblock {Reason2Drive}: Towards interpretable and chain-based reasoning for autonomous driving.
\newblock \emph{arXiv:2312.03661}, 2023.

\bibitem[Park et~al.(2022)Park, Xu, Yang, Keutzer, Kitani, Tomizuka, and Zhan]{park2022time}
Jinhyung Park, Chenfeng Xu, Shijia Yang, Kurt Keutzer, Kris Kitani, Masayoshi Tomizuka, and Wei Zhan.
\newblock Time will tell: New outlooks and a baseline for temporal multi-view 3d object detection.
\newblock \emph{arXiv:2210.02443}, 2022.

\bibitem[Philion and Fidler(2020)]{philion2020lift}
Jonah Philion and Sanja Fidler.
\newblock Lift, splat, shoot: Encoding images from arbitrary camera rigs by implicitly unprojecting to 3d.
\newblock In \emph{ECCV}, 2020.

\bibitem[Prakash et~al.(2021)Prakash, Chitta, and Geiger]{Prakash2021CVPR}
Aditya Prakash, Kashyap Chitta, and Andreas Geiger.
\newblock Multi-modal fusion transformer for end-to-end autonomous driving.
\newblock In \emph{CVPR}, 2021.

\bibitem[Qian et~al.(2023)Qian, Chen, Zhuo, Jiao, and Jiang]{qian2023nuscenes}
Tianwen Qian, Jingjing Chen, Linhai Zhuo, Yang Jiao, and Yu-Gang Jiang.
\newblock {NuScenes-QA}: A multi-modal visual question answering benchmark for autonomous driving scenario.
\newblock \emph{arXiv:2305.14836}, 2023.

\bibitem[Radford et~al.(2021)Radford, Kim, Hallacy, Ramesh, Goh, Agarwal, Sastry, Askell, Mishkin, Clark, et~al.]{radford2021learning}
Alec Radford, Jong~Wook Kim, Chris Hallacy, Aditya Ramesh, Gabriel Goh, Sandhini Agarwal, Girish Sastry, Amanda Askell, Pamela Mishkin, Jack Clark, et~al.
\newblock Learning transferable visual models from natural language supervision.
\newblock In \emph{ICML}, 2021.

\bibitem[Sachdeva et~al.(2024)Sachdeva, Agarwal, Chundi, Roelofs, Li, Kochenderfer, Choi, and Dariush]{sachdeva2024rank2tell}
Enna Sachdeva, Nakul Agarwal, Suhas Chundi, Sean Roelofs, Jiachen Li, Mykel Kochenderfer, Chiho Choi, and Behzad Dariush.
\newblock {Rank2Tell}: A multimodal driving dataset for joint importance ranking and reasoning.
\newblock In \emph{WACV}, 2024.

\bibitem[Sima et~al.(2023)Sima, Renz, Chitta, Chen, Zhang, Xie, Luo, Geiger, and Li]{sima2023drivelm}
Chonghao Sima, Katrin Renz, Kashyap Chitta, Li Chen, Hanxue Zhang, Chengen Xie, Ping Luo, Andreas Geiger, and Hongyang Li.
\newblock {DriveLM}: Driving with graph visual question answering.
\newblock \emph{arXiv:2312.14150}, 2023.

\bibitem[Tian et~al.(2024)Tian, Gu, Li, Liu, Hu, Wang, Zhan, Jia, Lang, and Zhao]{tian2024drivevlm}
Xiaoyu Tian, Junru Gu, Bailin Li, Yicheng Liu, Chenxu Hu, Yang Wang, Kun Zhan, Peng Jia, Xianpeng Lang, and Hang Zhao.
\newblock {DriveVLM}: The convergence of autonomous driving and large vision-language models.
\newblock \emph{arXiv:2402.12289}, 2024.

\bibitem[Touvron et~al.(2023)Touvron, Martin, Stone, Albert, Almahairi, Babaei, Bashlykov, Batra, Bhargava, Bhosale, et~al.]{touvron2023llama}
Hugo Touvron, Louis Martin, Kevin Stone, Peter Albert, Amjad Almahairi, Yasmine Babaei, Nikolay Bashlykov, Soumya Batra, Prajjwal Bhargava, Shruti Bhosale, et~al.
\newblock Llama 2: Open foundation and fine-tuned chat models.
\newblock \emph{arXiv:2307.09288}, 2023.

\bibitem[Vedantam et~al.(2015)Vedantam, Lawrence~Zitnick, and Parikh]{vedantam2015cider}
Ramakrishna Vedantam, C Lawrence~Zitnick, and Devi Parikh.
\newblock {CIDEr}: Consensus-based image description evaluation.
\newblock In \emph{CVPR}, 2015.

\bibitem[Wang et~al.(2024)Wang, Li, Li, Chen, Sima, Liu, Wang, Jia, Wang, Jiang, et~al.]{wang2024openlane}
Huijie Wang, Tianyu Li, Yang Li, Li Chen, Chonghao Sima, Zhenbo Liu, Bangjun Wang, Peijin Jia, Yuting Wang, Shengyin Jiang, et~al.
\newblock {OpenLane-V2}: A topology reasoning benchmark for unified 3d hd mapping.
\newblock \emph{NeurIPs}, 2024.

\bibitem[Wang et~al.(2023{\natexlab{a}})Wang, Liu, Wang, Li, and Zhang]{wang2023exploring}
Shihao Wang, Yingfei Liu, Tiancai Wang, Ying Li, and Xiangyu Zhang.
\newblock Exploring object-centric temporal modeling for efficient multi-view 3d object detection.
\newblock \emph{arXiv:2303.11926}, 2023{\natexlab{a}}.

\bibitem[Wang et~al.(2023{\natexlab{b}})Wang, Maalouf, Xiao, Ban, Amini, Rosman, Karaman, and Rus]{wang2023drive}
Tsun-Hsuan Wang, Alaa Maalouf, Wei Xiao, Yutong Ban, Alexander Amini, Guy Rosman, Sertac Karaman, and Daniela Rus.
\newblock {Drive Anywhere}: Generalizable end-to-end autonomous driving with multi-modal foundation models.
\newblock \emph{arXiv:2310.17642}, 2023{\natexlab{b}}.

\bibitem[Wang et~al.(2023{\natexlab{c}})Wang, Xie, Hu, Zou, Fan, Tong, Wen, Wu, Deng, Li, et~al.]{wang2023drivemlm}
Wenhai Wang, Jiangwei Xie, ChuanYang Hu, Haoming Zou, Jianan Fan, Wenwen Tong, Yang Wen, Silei Wu, Hanming Deng, Zhiqi Li, et~al.
\newblock {DriveMLM}: Aligning multi-modal large language models with behavioral planning states for autonomous driving.
\newblock \emph{arXiv:2312.09245}, 2023{\natexlab{c}}.

\bibitem[Wang et~al.(2022)Wang, Vitor~Campagnolo, Zhang, Zhao, and Solomon]{wang2022detr3d}
Yue Wang, Guizilini Vitor~Campagnolo, Tianyuan Zhang, Hang Zhao, and Justin Solomon.
\newblock {DETR3D}: 3d object detection from multi-view images via 3d-to-2d queries.
\newblock In \emph{CoRL}, 2022.

\bibitem[Wei et~al.(2022)Wei, Wang, Schuurmans, Bosma, Xia, Chi, Le, Zhou, et~al.]{wei2022chain}
Jason Wei, Xuezhi Wang, Dale Schuurmans, Maarten Bosma, Fei Xia, Ed Chi, Quoc~V Le, Denny Zhou, et~al.
\newblock Chain-of-thought prompting elicits reasoning in large language models.
\newblock \emph{NeurIPS}, 2022.

\bibitem[Wu et~al.(2023)Wu, Han, Wang, Liu, Zhang, and Shen]{wu2023language}
Dongming Wu, Wencheng Han, Tiancai Wang, Yingfei Liu, Xiangyu Zhang, and Jianbing Shen.
\newblock Language prompt for autonomous driving.
\newblock \emph{arXiv:2309.04379}, 2023.

\bibitem[Wu et~al.(2022)Wu, Jia, Chen, Yan, Li, and Qiao]{wu2022trajectoryguided}
Penghao Wu, Xiaosong Jia, Li Chen, Junchi Yan, Hongyang Li, and Yu Qiao.
\newblock Trajectory-guided control prediction for end-to-end autonomous driving: A simple yet strong baseline.
\newblock In \emph{NeurIPS}, 2022.

\bibitem[Xu et~al.(2023)Xu, Zhang, Xie, Zhao, Guo, Wong, Li, and Zhao]{xu2023drivegpt4}
Zhenhua Xu, Yujia Zhang, Enze Xie, Zhen Zhao, Yong Guo, Kenneth~KY Wong, Zhenguo Li, and Hengshuang Zhao.
\newblock {DriveGPT4}: Interpretable end-to-end autonomous driving via large language model.
\newblock \emph{arXiv:2310.01412}, 2023.

\bibitem[Zhai et~al.(2023)Zhai, Feng, Du, Mao, Liu, Tan, Zhang, Ye, and Wang]{zhai2023rethinking}
Jiang-Tian Zhai, Ze Feng, Jinhao Du, Yongqiang Mao, Jiang-Jiang Liu, Zichang Tan, Yifu Zhang, Xiaoqing Ye, and Jingdong Wang.
\newblock Rethinking the open-loop evaluation of end-to-end autonomous driving in nuscenes.
\newblock \emph{arXiv:2305.10430}, 2023.

\bibitem[Zhang et~al.(2021)Zhang, Liniger, Dai, Yu, and Van~Gool]{zhang2021roach}
Zhejun Zhang, Alexander Liniger, Dengxin Dai, Fisher Yu, and Luc Van~Gool.
\newblock End-to-end urban driving by imitating a reinforcement learning coach.
\newblock In \emph{ICCV}, 2021.

\bibitem[Zitkovich et~al.(2023)Zitkovich, Yu, Xu, Xu, Xiao, Xia, Wu, Wohlhart, Welker, Wahid, et~al.]{zitkovich2023rt}
Brianna Zitkovich, Tianhe Yu, Sichun Xu, Peng Xu, Ted Xiao, Fei Xia, Jialin Wu, Paul Wohlhart, Stefan Welker, Ayzaan Wahid, et~al.
\newblock {RT-2}: Vision-language-action models transfer web knowledge to robotic control.
\newblock In \emph{CoRL}, 2023.

\end{thebibliography}
}

% WARNING: do not forget to delete the supplementary pages from your submission 
% \input{sec/X_suppl}

\end{document}